\documentclass[11pt]{article}

\usepackage{graphicx}

\usepackage{amsmath,amssymb} 
\usepackage{color}
\usepackage{subfigure}
\usepackage{enumerate}
\usepackage{algorithm}
\usepackage{algorithmic}
\usepackage{url}

\DeclareMathOperator*{\argmin}{\arg\min}

\usepackage{amsthm}

\theoremstyle{plain}
\newtheorem{prop}{Proposition}
\theoremstyle{definition}
\newtheorem{defn}{Definition}

\begin{document}

\title{Occlusion-Model Guided Anti-Occlusion Depth Estimation in Light Field} 

\author{Hao Zhu, Qing Wang\\
School of Computer Science\\
Northwestern Polytechnical University\\
qwang@nwpu.edu.cn
\and
Jingyi Yu\\
University of Delaware\\
jingyiyu@udel.edu
}

\date{}
\maketitle

\begin{abstract}
Occlusion is one of the most challenging problems in depth estimation. Previous work has modeled the single-occluder occlusion in light field and get good results, however it is still difficult to obtain accurate depth for multi-occluder occlusion. In this paper, we explore the multi-occluder occlusion model in light field, and derive the occluder-consistency between the spatial and angular space which is used as a guidance to select the un-occluded views for each candidate occlusion point. Then an anti-occlusion energy function is built to regularize depth map. The experimental results on public light field datasets have demonstrated the advantages of the proposed algorithm compared with other state-of-the-art light field depth estimation algorithms, especially in multi-occluder areas.
\end{abstract}

\section{Introduction}
Depth estimation has been researched for decades. A common and important assumption is the photo-consistency assumption, \textit{i.e.}, the colors of a point observed from different views ought to be similar. This assumption holds for non-occlusion points. However, it fails for occlusion points as these points can not be observed from all views.

Many works have been done to handle occlusions. Kolmogorov \textit{et al.} \cite{kolmogorov2002multi} encodes the visibility constraint and introduces an occlusion term to smooth it. Woodford \textit{et al.} \cite{woodford2009global} adds an additional second order smoothness terms and use Quadratic Pseudo-Boolean Optimization to solve it. Then, Bleyer \textit{et al.} \cite{rother2007optimizing} applies the asymmetric occlusion model to improve depth estimation. However, due to the wide baseline and the lack of views, these works only notice the image of the occlusion, \textit{i.e.}, the occlusion point is visible in reference view and invisible in other views, and heavy occlusion can not be well handled.

Light-field cameras from Lytro \cite{lytro_web} and Raytrix \cite{raytrix_web} obtain a 4D light field by inserting a micro-lens array into the traditional camera \cite{ng2005light}, which provides hundreds of views in a single shot. Apart from this, the baseline between these views is very small, which means no aliasing occurs and the consistent correspondences still holds in this case \cite{xiao2014aliasing}. Combining the multi-view and the micro-baseline, a more detailed and complete representation of occlusion appears, which renews the method to handle occlusion.

Wanner \textit{et al.} \cite{wanner2012globally,wanner2014variational} applies structure tensor to analyze the Epipolar Plane Image (EPI). This method only takes advantage of angular samples in one dimension, and the tensor field becomes too random to analyze in heavy occlusion. Yu \textit{et al.} \cite{yu2013line} encodes the constraints of 3D lines and introduces Lind Assisted Graph Cuts (LAGC) to improve depth estimation. However, the 3D lines are partitioned into small and incoherent segments in heavy occlusion which leads to wrong estimation. Chen \textit{et al.} \cite{chen2014light} proposes to select the un-occluded views by using a bilateral metric in angular space. However, this selection lacks the guidance of the physical model and the number of un-occluded views is also a predefined parameter. Wang \textit{et al.} \cite{wang2015occlusion,wang2016depth} analyzes the formation of occlusion and finds the consistency between the spatial patch and the angular patch in occlusion boundaries. They select the un-occluded views according to the edges in spatial patch. However, their method failed in multi-occluder areas where the local patch can not be divided into two regions by a straight line, and it leads to over smooth results in these areas (Fig. \ref{fig:depth_map1}, \ref{fig:depth_map2}). Although Wang \textit{et al.} have a more recent work \cite{wang2016svbrdf} for depth estimation, it aims at solving shadings which is not applicable in our case.

In this paper, we explore the light field occlusion theory for multi-occluder occlusion, and propose an algorithm to regularize depth map. Our main contributions are:
\begin{enumerate}[1)]
\item The light field occlusion theory for multi-occluder occlusions.
\item An algorithm to accurately select the un-occluded views in angular space with the guidance of the occlusion theory which does not need any predefined parameters.
\item A depth estimation algorithm which is robust to multi-occluder occlusion.
\end{enumerate}

In Section 2, we model the multi-occluder occlusion in light field, and derive the occlusion-consistency between the spatial and angular space, \textit{i.e.}, the occluded views in angular space are projections of the occluder in spatial space. With the guidance of occluder-consistency, we select the un-occluded views for each candidate occlusion point using a partition in spatial patch in Section \ref{sec:OAS:inioas}, and obtain an initial depth map using the un-occluded views in Section \ref{sec:OAS:inidepest}. Then, we detect the occlusion points from the initial depth map according to the visible constraint in Section \ref{sec:DR:occdet}. Finally the occlusion map and the un-occluded views are used to build an anti-occlusion energy function to refine depth in Section \ref{sec:DR:fdr}. In Section \ref{sec:ExpRe}, we provide complete comparisons with other state-of-the-art algorithms, both in quantitative and in qualitative, showing great advantages compared with previous works \cite{wanner2012globally,wanner2014variational,yu2013line,chen2014light,wang2015occlusion}. 

\section{The Light Field Occlusion Model}
\label{sec:Occ_model}

In this section, we will analyze the formation of the occlusion based on the physical model, and explore the occlusion theory for multi-occluder occlusion. 

\subsection{Definitions and Notations}
Before building the light field occlusion model, we first give some definitions and notations of the light field model. 
\begin{defn}
\label{defn:single_occ}
	Single-occluder occlusion refers to 	the occlusion that occluded views and un-occluded views can be divided half-and-half.
\end{defn}
\begin{defn}
\label{defn:multi_occ}
	Multi-occluder occlusion refers to the occlusion that there are more occluded views than un-occluded views.
\end{defn}

Point \((x,y,u,v)\) in 4D light field describes a ray emitting from the world point \((X,Y,Z)\), and intersecting with two parallel planes, namely, the \(uv\) camera plane and the \(xy\) image plane, which we refer to as angular and spatial space. Tab. \ref{tab:notations} presents a list of terms used throughout this section. 

\setlength{\tabcolsep}{4pt}
\begin{table}
\begin{center}
\caption{
Notations of the light field occlusion model.
}
\label{tab:notations}
\begin{tabular}{ll}
\hline\noalign{\smallskip}
Term & Definition\\
\noalign{\smallskip}
\hline
\noalign{\smallskip}
\((u,v)\) & angular/camera plane coordinates\\
\((x,y)\) & spatial/image plane coordinates\\
\((X,Y,Z)\) & world coordinates\\
\(\vec{e}_A\) & the vector in angular coordinate system\\
\(\vec{e}_I\) & the vector in spatial coordinate system\\
\(\vec{e}_W\) & the vector in world coordinate system\\
\hline
\end{tabular}
\end{center}
\end{table}
\setlength{\tabcolsep}{1.4pt}

\begin{figure}[t]
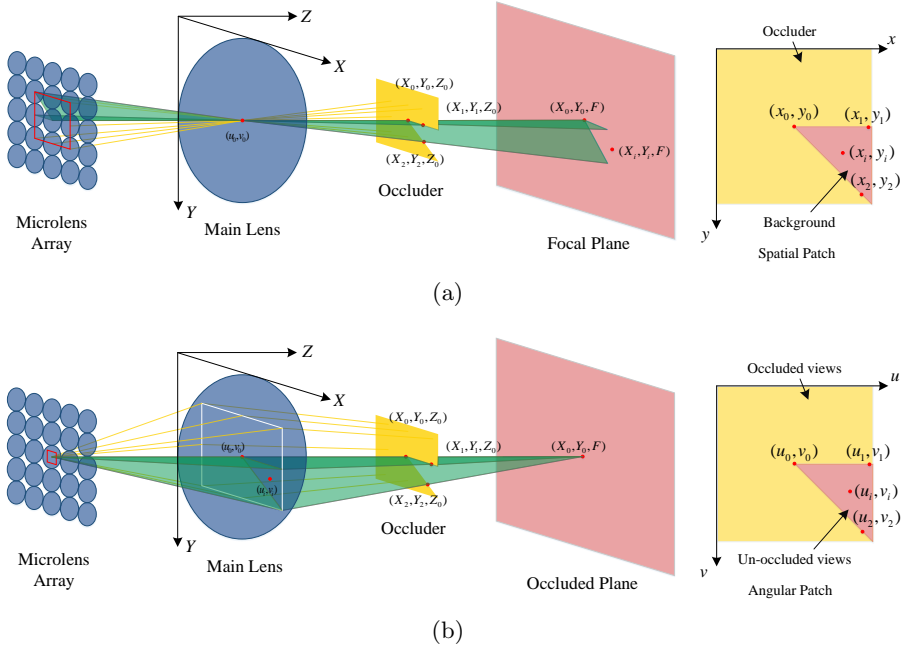

\begin{center}
\centering
\subfigure[]{
	\label{fig:pc_uoc:pinhole}
	\includegraphics[width=120mm]{Occlusion_model_pinhole} 
}
\subfigure[]{
\label{fig:pc_uoc:scam}
	\includegraphics[width=120mm]{Occlusion_model_scam}
}
\end{center}
\caption{The light field camera model with occlusion. The left image in (a) denotes the the image captured from the view \((u_{0},v_{0})\), and the right image in (a) is a local patch centered at \((x_{0},y_{0})\) from the view \((u_{0},v_{0})\). The left image in (b) denotes the light field is refocused in depth \(F\), and only views constrained by two green planes can see the point \((X_{0},Y_{0},F)\), the images formed from other views describe the occluder. The right image in (b) is the angular patch of the point \((x_{0},y_{0})\).}
\label{fig:pc_uoc}
\end{figure}

\subsection{Occlusion Model}
Previous work \cite{wang2015occlusion} has proved the occluder-consistency for single-occluder occlusion, \textit{i.e.}, when refocused to the correct depth, the edge which separates the un-occluded and occluded views in the angular patch has the same orientation as the occlusion edge in spatial patch. This property is useful for single-occluder occlusions, however it fails in multi-occluder situation as the un-occluded and occluded pixels in angular patch can not be divided into two regions by a straight line. 

We first consider a simple multi-occluder occlusion (Fig. \ref{fig:pc_uoc}). Considering a pixel \((X_0,Y_0,F)\) on the focal plane (the left image in Fig. \ref{fig:pc_uoc:pinhole}), and an occluder intersects at \((X_0,Y_0,Z_0)\) (\(0<Z_0<F\)). Note that the occluder has two edges, and the directional vectors of these two edges in the plane \(Z=Z_0\) are,
\begin{equation}
\begin{aligned}
	\vec{e}_W^1 & = (k_{X_1},k_{Y_1}) = (X_1-X_0,Y_1-Y_0)\\
	\vec{e}_W^2 & = (k_{X_2},k_{Y_2}) = (X_2-X_0,Y_2-Y_0).
\end{aligned}
\end{equation}
The larger angle between these two vectors denotes occluded areas (the golden areas in Fig. \ref{fig:pc_uoc}). Without loss of generality, we assume \(k_{Y_{j}}>0,j=1,2\). 

For any other pixel \((X_{i},Y_{i},F)\) on the focal plane, it will be observed by the view \((u_0,v_0)\) iff it meets the following inequalities,
\begin{equation}
\label{eqn:occ_cons_spatial}
	\begin{aligned}
		&k_{Y_1}(X_i-X_{0})-k_{X_1}(Y_i-Y_{0}) > 0\\
		&k_{Y_2}(X_i-X_{0})-k_{X_2}(Y_i-Y_{0}) < 0.
	\end{aligned}
\end{equation}

We then project these inequalities from world coordinate system to the image system (the right image in the Fig. \ref{fig:pc_uoc:pinhole}). The corresponding directional vectors of the \(\vec{e}_W^1\) and \(\vec{e}_W^2\) are $\vec{e}_I^1=(k_{x_1},k_{y_1})$ and $\vec{e}_I^2=(k_{x_2},k_{y_2})$ respectively. $\vec{e}_I^1=\lambda_1 \vec{e}_W^1$ and $\vec{e}_I^2=\lambda_1 \vec{e}_W^2$, \(\lambda_{1}\) is a scale factor to denote the scaling relationship between the world coordinate system and the image coordinate system. For any other point \((x_i,y_i)\) on the image, it is a background point iff,
\begin{equation}
\label{eqn:occ_cons_spatial_img}
	\begin{aligned}
		&k_{y_1}(x_i-x_{0})-k_{x_1}(y_i-y_{0}) > 0\\
		&k_{y_2}(x_i-x_{0})-k_{x_2}(y_i-y_{0}) < 0.
	\end{aligned}
\end{equation}

Then considering the main lens plane (the left image in Fig. \ref{fig:pc_uoc:scam}. The light field is refocused to the depth \(F\)). For any other view \((u_{i},v_{i})\) on the main lens plane, it can capture the pixel \((X_0,Y_0,F)\) iff
\begin{equation}
\label{eqn:occ_cons_angular}
	\begin{aligned}
		&k_{v_1}(u_i-u_{0})-k_{u_1}(v_i-v_{0}) > 0\\
		&k_{v_2}(u_i-u_{0})-k_{u_2}(v_i-v_{0}) < 0.
	\end{aligned}
\end{equation} 
where $\vec{e}_A^j=(k_{u_j},k_{v_j})$, $\vec{e}_A^j = \lambda_2\vec{e}_I^j$, $j=1,2$, \(\lambda_{2}\) is a scale factor to denote the scaling relationship between the image coordinate system and the angular coordinate system. 

Revisiting the Eqn. \ref{eqn:occ_cons_spatial_img} and \ref{eqn:occ_cons_angular}, it is noticed Eqn. \ref{eqn:occ_cons_spatial_img}, \ref{eqn:occ_cons_angular} have the same inequalities and \((u,v)\), \((x,y)\) are one-one correspondence, as \((u_0,v_0)\) and \(x_0,y_0\) are on a same line, the following proposition can be obtained, 
\begin{prop}
\label{prop_project}
	The occluded views in angular space are projection of the occluder in spatial space.
\end{prop}
In other words, in a local spatial patch, the corresponding views of the occluder are the occluded views, and the corresponding views of the background are the un-occluded views. This proposition is called occluder-consistency later.

Note that, the proposition mentioned above is obtained under a simple multi-occluder assumption. For a more complex multi-occluder, the boundaries of the occluder can be divided into more small straight lines by following the idea of the Calculus, and the Eqn. \ref{eqn:occ_cons_spatial_img}, \ref{eqn:occ_cons_angular} will contain more inequalities. No matter how many inequalities, the numbers of inequalities in Eqn. \ref{eqn:occ_cons_spatial_img}, \ref{eqn:occ_cons_angular} are equal and inequalities are one-one correspondence, and the Prop. \ref{prop_project} always holds.

\subsection{Projection Radius}
\begin{figure}[tbp]
\begin{center}
\centering
\subfigure[]{
	\label{fig:occ_projection_radius_b}
	\includegraphics[width=35mm]{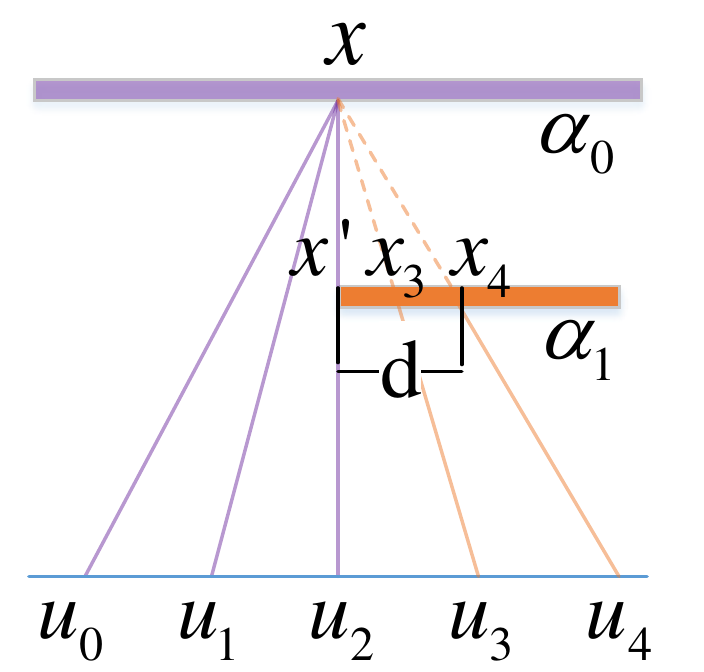}
}
\subfigure[]{
	\label{fig:occ_projection_radius_c}
	\includegraphics[width=35mm]{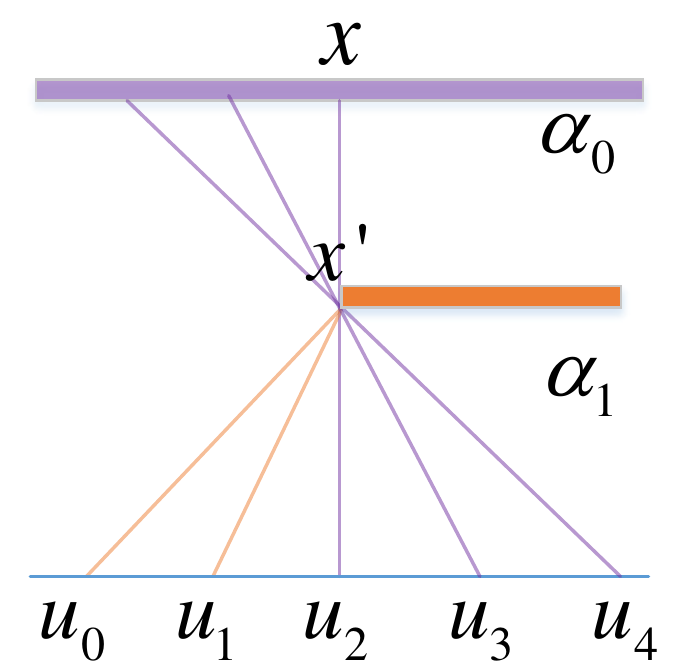}
}
\end{center}
\caption{Projection radius. In (a), the light field is refocused to the background in depth \(\alpha_0\), where the light from \(u_0, u_1, u_2\) converge to point \(x\) and the light from \(u_3, u_4\) are blocked by the occluder. In (b), the light field is refocused to the foreground in depth \(\alpha_1\), and light from all views converge to point \(x'\).} 
\label{fig:occ_projection_radius}
\end{figure}

For occluded points, the occluded views in angular space are projection of the occluder in spatial space in a local patch. We derive the radius of the patch in a 2D light field. In Fig. \ref{fig:occ_projection_radius}, the purple lines denote the background at depth \(\alpha_0\), the orange lines denote the occluder at depth \(\alpha_1\), blue lines denote the camera plane, \(x\) denotes a pixel in the background, and \(u_0-u_4\) denote different views in light field.

Firstly, the light field is refocused to the background at depth \(\alpha_0\) (Fig. \ref{fig:occ_projection_radius_b})\cite{ng2006digital},
\begin{equation}
	LF_{\alpha_0}(x,u) = LF_{0}(x+(u-u_2)(1-\frac{1}{\alpha_0}),u),
\end{equation}
where \(LF_0\) is the input light field, \(LF_{\alpha_0}\) is the refocused light field at depth \(\alpha_0\), \(u_2\) is the central view of the light field. It is noticed the light from views \(u_0, u_1, u_2\) converge to the point \(x_2\), and the light from \(u_3, u_4\) are blocked by the occluder. \(u_3, u_4\) are the occluded views in angular space, and the images of these two views come from points \(x_3, x_4\). In other words, the horizontal distance \(d\) between \(x_4\) and \(x\) is the projection radius.

Then, the light field is refocused to the foreground at depth \(\alpha_1\) (Fig. \ref{fig:occ_projection_radius_c})
\begin{equation}
	LF_{\alpha_1}(x,u) = LF_{0}(x+(u-u_2)(1-\frac{1}{\alpha_1}),u),
\end{equation}
where \(LF_{\alpha_1}\) is the refocused light field at depth \(\alpha_1\). It can be seen the light from all views converge to the point \(x'\). As the horizontal distance between \(x'\) and \(x_2\) is 0, the projection radius \(d\) is obtained
\begin{equation}
\begin{aligned}
d
&=|x+(u_4-u_2)(1-\frac{1}{\alpha_1})-x-(u_4-u_2)(1-\frac{1}{\alpha_0})|\\
&=|(u_4-u_2)(\frac{1}{\alpha_0}-\frac{1}{\alpha_1})|.
\end{aligned}
\label{eqn:projection_raidus}
\end{equation}

\section{Depth Initialization}
\label{sec:OAS}
In this section, we will show how to use the occluder-consistency between the spatial and angular space to select the un-occluded views, and how to obtain an initial depth estimation using these un-occluded views.

\begin{figure}[t]
\begin{center}
\centering
\subfigure[]{
	\includegraphics[width=38mm]{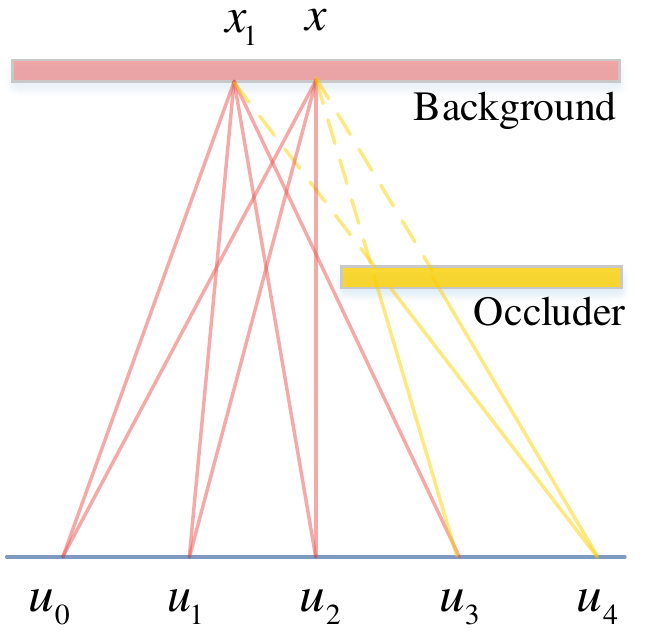} 
}
\subfigure[]{
	\includegraphics[width=27mm]{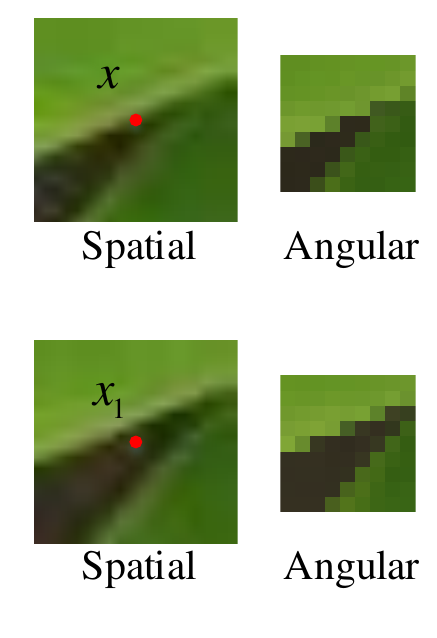}
}
\end{center}
\caption{The occlusion in different views. (a) demonstrates the model of occlusion in central view and in other views respectively. Point \(x\) is occluded in central view and point \(x_1\) is occluded in other views. (b) demonstrates the image of occlusion in central view and in other views respectively. It is noticed \(x\) is an edge point, and there are many occluded views in its angular patch when refocused to its true depth. The point \(x_1\) is not an edge point, however as it near the edges, there are many occluded views in its angular patch too.}
\label{fig:occ_in_dif_views}
\end{figure}

\subsection{Un-Occluded Views Selection}
\label{sec:OAS:inioas}
We first give an important assumption about occlusion of the proposed algorithm. The occluder has a different color of the occluded point. For the situation that the occluder is similar to occluded point, as far as we know, no work can handle it. Based on this assumption, the following proposition holds,
\begin{prop}
\label{prop_occ_assum}
	An occlusion point is an edge point but an edge point may not be an occlusion point.
\end{prop}

\begin{figure}[tbp]
\centering
	\includegraphics[width=80mm]{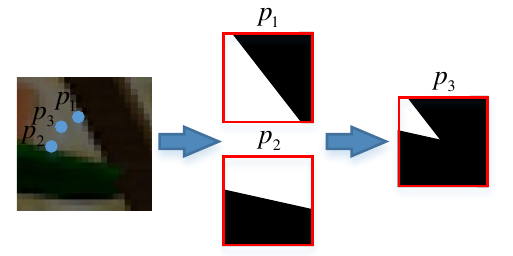}
\caption{The voting strategy of obtaining un-occluded views for pixels occluded in other views. The first col shows that points \(p_1, p_2\) are occluded in central view and \(p_3\) is occluded in other views. The second row shows the un-occluded views of \(p_1, p_2\) (the areas labeled with white are un-occluded views). The third col is the un-occluded views of \(p_3\) after voting.}
\label{fig:unocc_dilate}
\end{figure}

With Prop. \ref{prop_occ_assum}, the canny edge detector is firstly applied to find the candidate occlusion points \(S_{occ}\). Then the K-means clustering \cite{anzai2012pattern} is applied for the local image patch\footnote{The patch size is set as half of the angular resolution of light field initially since we do not have depth map here.} centered at each occlusion point \(p\) from \(S_{occ}\) (the feature is the RGB color, and the number of labels is 2). For each patch, the pixels which share the same label with the center pixel are labeled as background or un-occluded points. According to the occluder-consistency mentioned in Prop. \ref{prop_project}, the corresponding views in angular patch of the center pixel \(p\) are labeled as un-occluded views \(\Omega_{p}^{uocc}\). These pixels are candidate occlusion pixels in the central view (Fig. \ref{fig:occ_in_dif_views}). For pixels occluded in other views, the un-occluded views are voted from its neighborhood system (Fig. \ref{fig:unocc_dilate}). 

\subsection{Depth Estimation}
\label{sec:OAS:inidepest}
With the un-occluded views selection, a robust initial depth estimation is obtained based on the classical photo-consistency in un-occluded views. 

We refocus the light field to different depth \(\alpha\),
\begin{equation}
	LF_{\alpha}(x,y,u,v) = LF_{0}(x+u(1-\frac{1}{\alpha}),y+v(1-\frac{1}{\alpha}),u,v),
\end{equation}
where \(LF_{0}\) is the input 4D light field, \(LF_{\alpha}\) is the refocused light field in depth \(\alpha\), \((x,y)\) are the spatial coordinates, and \((u,v)\) are the angular coordinates. Then, the matching cost of each pixel \(p\) is defined as,
\begin{equation}
	C_{\alpha}^{uocc}(p) = \frac{1}{|\Omega_{p}^{uocc}|}\sum_{(u,v)\in \Omega_{p}^{uocc}}\left|A_{p_{\alpha}}(u,v)-A_{p_{\alpha}}(0,0)\right|,
\label{eqn:uocc}
\end{equation}
where \(\Omega_{p}^{uocc}\) is the un-occluded views set of the point \(p\) (Sec. \ref{sec:OAS:inioas}), and \(|\cdot|\) denotes the size of the set \(\cdot\), and \(A_{p_{\alpha}}(u,v)\) denotes the angular image of pixel \(p\) at depth \(\alpha\) (In other words, \(A_{p_\alpha}(u,v) = LF_{\alpha}(x_p,y_p,u,v)\), \((x_p,y_p)\) are the spatial coordinates of pixel \(p\)), and \(A_{p_{\alpha}}(0,0)\) is the color of pixel \(p\) in central view.

Then, the initial depth estimation of each pixel \(p\) is obtained,
\begin{equation}
	\alpha_{ini}(p) = \argmin_{\alpha}C_{\alpha}^{uocc}(p).
\end{equation}

\section{Depth Regularization}
\label{sec:DepthRegu}
In this section, we will show how to find the occlusion and regularize it with a global energy function.

\subsection{Occlusion Detection}
\label{sec:DR:occdet}
We find the occlusion point using the visible constraint, \textit{i.e.}, the occlusion point is visible in reference view and invisible in other views. In other words, if the difference of disparities of two neighboring pixels is larger than 1 pixel, there is an occlusion point here. In light field, this constraint can only find the occlusion point in the central view due to the multiple views, and the threshold value \(\epsilon_{occ}\) of the difference of the disparities ought to be relaxed to find the occlusion in other views,
\begin{equation}
	\epsilon_{occ} = \frac{1}{\lfloor N_{uv}/2\rfloor },
\end{equation}
where \(N_{uv}\) is the angular resolution \footnote{The angular resolution of the light field that we use in the experiment is \(9\times 9\).}.

As the initial depth estimation in occluded points is unreliable, and sometimes it is too smooth and random to distinguish the occlusion point accurately from only two neighboring pixels. We select a disparity patch centered at each candidate occlusion point (the detected points by the Canny operator), and use the K-means clustering to divide the patch into 2 classes, then the difference of disparities of each class \(\Delta dis\) is determined by the subtraction of two centers,
\begin{equation}
	\Delta dis = |dis_{1} - dis_{2}|,
\end{equation}
where \(dis_{i}\) is the center of the \(i\)-th class.

Finally the candidate occlusion point \(p\) is determined by comparing the \(\epsilon_{occ}\) and \(\Delta dis(p)\),
\begin{equation}
	Occ(p)=
	\begin{cases}
		1 & \Delta dis(p) \geqslant \epsilon_{occ}\\
		0 & \Delta dis(p) < \epsilon_{occ}.
	\end{cases}
\end{equation}

\subsection{Un-occluded Views Re-Selection}
\label{sec:DR:re_oas}
For each occlusion point \(p\), we apply the K-means clustering for its local depth patch to find the background depth \(\alpha_0\) and occluder depth \(\alpha_1\), and the projection radius \(d\) is determined using Eqn. \ref{eqn:projection_raidus}. Then, each patch is resized to the angular resolution of light field. The following procedure is the same as Sec. \ref{sec:OAS:inioas}. 

Since the occlusion points can not be fully detected in Sec. \ref{sec:DR:occdet}, we retain the un-occluded views selection for other candidate points obtained in Sec. \ref{sec:OAS:inioas}.
\subsection{Final Depth Regularization}
\label{sec:DR:fdr}
Finally, given the occlusion cues, we regularize with Markov Random Field (MRF) for a final depth map.
\begin{equation}
\label{eqn:DR:energyfunction}
	E = \sum_{p}E_{uocc}(p,\alpha_{p})+\lambda_{s}\sum_{p,q}E_{smooth}(p,q,\alpha_{p},\alpha_{q})
\end{equation}
where \(\alpha_{p}\) is the depth of pixel \(p\), and \(p, q\) are neighboring pixels, and \(\lambda_{s}\)(=0.35 in our experiment) is to control the smooth term. 

The data term \(E_{uocc}\) measures the photo-consistency in un-occluded views, 
\begin{equation}
	E_{uocc}(p,\alpha_{p}) = 1-e^{-\frac{C_{\alpha_{p}}^{uocc}(p)^2}{2\sigma_{uocc}^{2}}},
\end{equation}
where \(\sigma_{uocc}\)(=3 in our experiment) controls the sensitivity of the function to large distance, and the definition of \(C_{\alpha_{p}}^{uocc}(p)\) can be found in Eqn. \ref{eqn:uocc}.

The smooth term \(E_{smooth}\) encodes the smoothness constraint between two neighboring pixels,
\begin{equation}
\begin{aligned}
	E_{smooth}(p,q,\alpha_{p},\alpha_{q})
	&=\omega_{pq}|\alpha_{p}-\alpha_{q}|,\\
	\omega_{pq}
	&=e^{-\frac{(Occ(p)-Occ(q))^2}{2\gamma_{occ}^2}-\frac{(I_e(p)-I_e(q))^2}{2\gamma_{e}^2}-\frac{(I(p)-I(q))^2}{2\gamma_{c}^2}},
\end{aligned}
\end{equation}
where \(I_e\) is the edge map of the central view image \(I\), and \((\gamma_{occ}, \gamma_{e}, \gamma_{c})\) are three weighting factors. Comparing with previous works, we introduce the occlusion term and the edge term into the weighting function \(\omega_{pq}\) to preserve occlusion boundaries and keep the depth of occlusion boundaries similar. 

The final depth map is obtained by minimizing the Eqn. \ref{eqn:DR:energyfunction}\cite{boykov2004experimental,boykov2001fast,kolmogorov2004energy},
\begin{equation}
	\alpha^{*} = \argmin_{\alpha}E.
\end{equation}

The full description of the proposed algorithm is given in Algo. \ref{algo-1}. First, edge detection is applied on central view image \(I\) to find all possible occlusion point \(I_e\). Then, the un-occluded views \(\Omega^{uocc} \) of each candidate occlusion point are selected based on a K-means clustering. After that, an initial depth map \(\alpha_{ini}\) is estimated by using the un-occluded views. Moreover, the occlusion map \(Occ\) is detected by using the initial estimation. Finally, based on the un-occluded views and the occlusion map, the depth map \(\alpha_{final}\) is regularized with a MRF energy function.

\begin{algorithm}
	\begin{algorithmic}
		\STATE \textbf{Input:}\\
		\STATE 4D light field \(LF\)
		
		\STATE \textbf{Output:}\\ 
		\STATE Final depth map \(\alpha_{final}\)
				
		\STATE \textbf{Process:} \\
		\STATE \(I_e \quad\ \ \;\,= canny(I)\)\hfill \(\triangleright\) Sec. \ref{sec:OAS:inioas} \\
		\STATE \(\Omega^{uocc} \ \,= kmeans(I_e, I)\)\hfill \(\triangleright\) Sec. \ref{sec:OAS:inioas} \\
		\STATE \(\alpha_{ini} \quad\,= InitialDepthEst(LF,I_e,\Omega^{uocc})\)\hfill \(\triangleright\) Sec. \ref{sec:OAS:inidepest} \\
		\STATE \(Occ \quad\;= OccDetection(\alpha_{ini}, I_e)\)\hfill \(\triangleright\) Sec. \ref{sec:DR:occdet} \\
		\STATE \(E \qquad\,= E_{uocc}(LF,\Omega^{uocc})+E_{smooth}(I,I_e,Occ)\)\hfill \(\triangleright\) Sec. \ref{sec:DR:fdr} \\
		\STATE \(\alpha^{*} \quad\;\;\ = \arg \min_{\alpha} E\)\hfill \(\triangleright\) Sec. \ref{sec:DR:fdr}
		
	\end{algorithmic}
	\caption{The proposed algorithm}
	\label{algo-1}
\end{algorithm}

\section{Experimental Results}
\label{sec:ExpRe}
We compare our results with the globally consistent depth labeling (GCDL) by Wanner \textit{et al.} \cite{wanner2012globally}, the line-assigned graph-cut(LAGC) by Yu \textit{et al.} \cite{yu2013line}, the Bilateral Consistency Metric (BCM) by Chen \textit{et al.} \cite{chen2014light} and the occlusion-aware depth estimation (OADE) by Wang \textit{et al.} \cite{wang2015occlusion}. Note that, the results of GCDL come from their published papers \cite{wanner2013datasets}, the results of LAGC and OADE are obtained by running their published codes or executable files, and the results of BCM are provided by the authors.

The performance of the proposed algorithm is evaluated by using the most popular light field datasets \cite{wanner2013datasets}. This datasets are synthesized by the Blender, and each data includes a \(9\times 9\) light field and its ground truth depth. For runtime, on a 3.4 GHz Intel i7 machine with 16 GB RAM, our MATLAB implementation takes about 1 hour on a \(9\times 9\times 768\times 768\) color light field. Considering the precise results and the low-speed of MATLAB, this time cost is acceptable.

\subsection{Un-occluded views selection}
\label{sec:ExpRe:uocs}
\begin{figure}[t]
\centering
	\includegraphics[width=120mm]{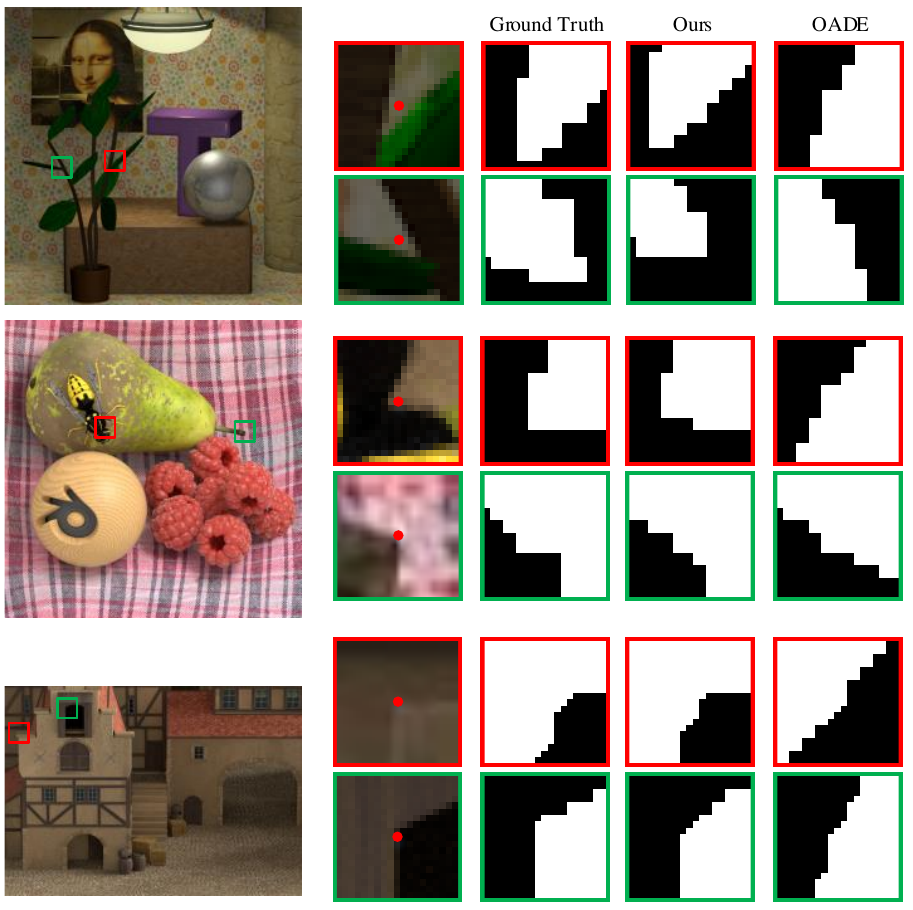}
\caption{The comparisons of the un-occluded views selection. The areas labeled with white are the selected un-occluded views.}
\label{fig:unocc_selection}
\end{figure}

A consensus on depth estimation in computer vision is that more effective views lead to more accurate depth estimation. So, the precision and the recall of the selected un-occluded views are important. We count the F-measure (the harmonic mean of precision and recall compared with the ground truth) of the un-occluded views in occlusion using our algorithm, and compare it with previous work \cite{wang2015occlusion}. The quantitative comparisons are listed in Tab. \ref{tab:oas_selection}, and the qualitative comparisons are shown in Fig. \ref{fig:unocc_selection}. It can be seen that our selection method outperforms previous work in the un-occluded views selection. And this advantage is more obvious especially in multi-occluder areas. In Fig. \ref{fig:unocc_selection}, our method can always select accurate un-occluded views, however, Wang's \textit{et al.} \cite{wang2015occlusion} method always selects more occluded views and these selections will lead to over smooth results in occlusion areas (Fig. \ref{fig:depth_map1}, \ref{fig:depth_map2}). It is noticed our method performs not good in Horses. The reason is that there are many textures near the occlusion boundaries in background, and the K-means clustering based on color can not divide the background and occluder accurately in the complex texture areas.

\setlength{\tabcolsep}{4pt}
\begin{table}
\begin{center}
\caption{
The F-measure of un-occluded views selection
}
\label{tab:oas_selection}
\begin{tabular}{cccccccc}
\hline\noalign{\smallskip}
									& Buddha &Buddha2 &Horses &Medieval &Mona &Papillon &StillLife\\
\noalign{\smallskip}
\hline
\noalign{\smallskip}
OADE\cite{wang2015occlusion}     	& \textbf{0.71} & 0.73 & \textbf{0.66} & 0.58 & 0.69 & 0.59 & 0.59\\
Our method 							& \textbf{0.71} & \textbf{0.74} & 0.62 & \textbf{0.60} & \textbf{0.79} & \textbf{0.79} & \textbf{0.72}\\
\hline
\end{tabular}
\end{center}
\end{table}
\setlength{\tabcolsep}{1.4pt}

\subsection{Occlusion Boundaries}
\label{sec:ExpRe:occb}
\begin{figure}[t]
\centering
	\includegraphics[width=120mm]{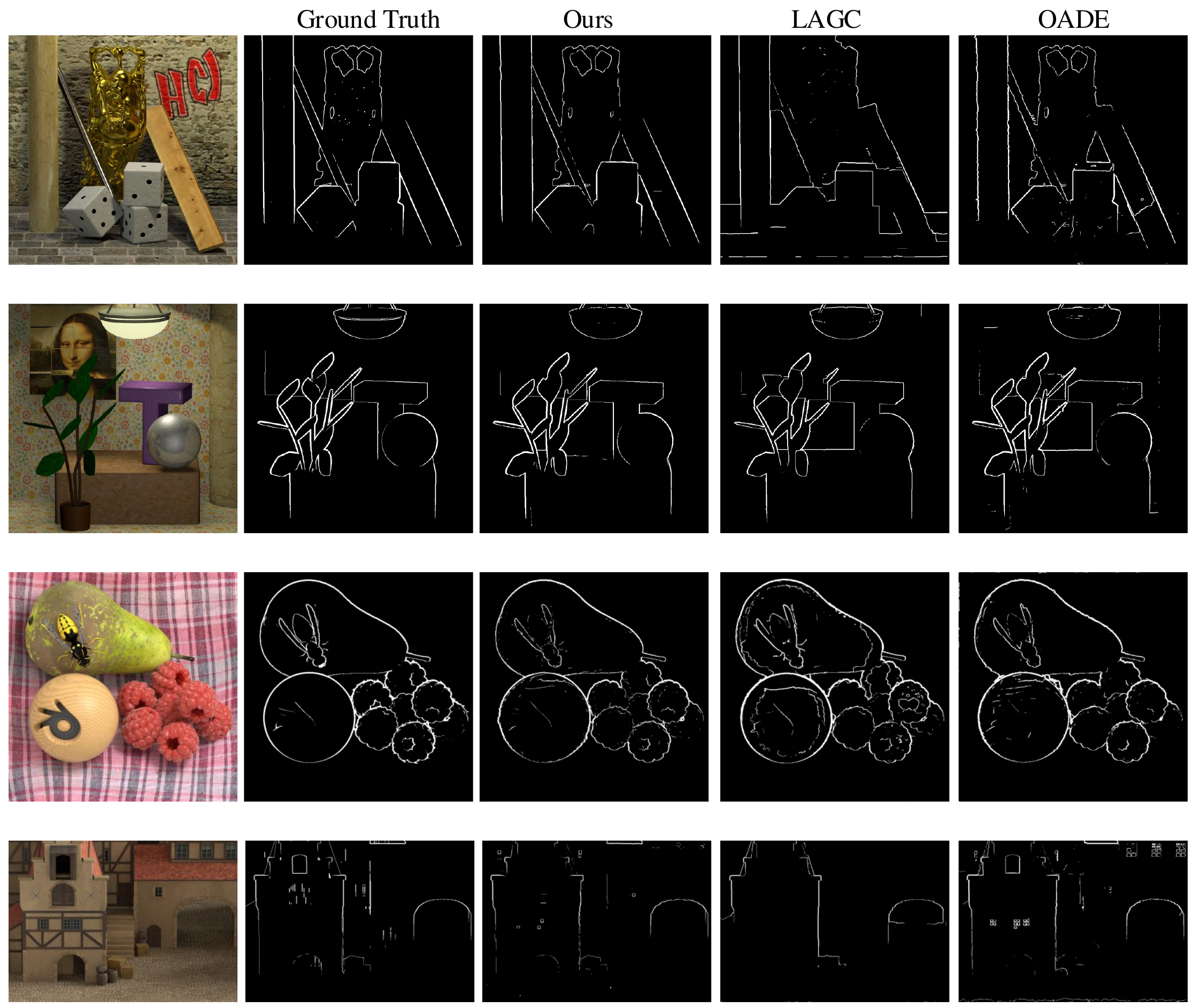}
\caption{The detected occlusion boundaries.}
\label{fig:occ_bound}
\end{figure}

For each data, we detect its occlusion boundaries using the depth map, and compute its F-measure. Then, we compare it with other state-of-the-art algorithms. The quantitative comparisons are listed in Tab. \ref{tab:occ_bound}, and the qualitative comparisons are shown in Fig. \ref{fig:occ_bound}. Our algorithm outperforms the previous works. Note that the results of GCDL \cite{wanner2012globally} and BCM \cite{chen2014light} are not contained as it is difficult to run their codes in our experimental environment. However, as previous works \cite{yu2013line,wang2015occlusion} have demonstrated their advantages to \cite{wanner2012globally} and \cite{chen2014light}, these comparisons are convincing. Our method performs not good in StillLife (the third row in Fig. \ref{fig:occ_bound}). That is because there are many weak occlusions (the difference of disparities is small) in StillLife. The difference of disparities in the boundaries of the bee is small, the occlusion detection method in Sec. \ref{sec:DR:occdet} can not handle these occlusions well.
\setlength{\tabcolsep}{4pt}
\begin{table}[!t]
\begin{center}
\caption{
The F-measure of occlusion boundaries.
}
\label{tab:occ_bound}
\begin{tabular}{cccccccc}
\hline\noalign{\smallskip}
									& Buddha &Buddha2 &Horses &Medieval &Mona &Papillon &StillLife\\
\noalign{\smallskip}
\hline
\noalign{\smallskip}
LAGC\cite{yu2013line} 				& 0.54 & 0.41 & 0.55 & 0.32 & 0.64 & 0.53 &	0.49\\
OADE\cite{wang2015occlusion}     	& 0.71 & 0.70 & 0.75 & 0.47 & 0.75 & 0.65 & \textbf{0.82}\\
Our method 							& \textbf{0.75} & \textbf{0.85} & \textbf{0.80} & \textbf{0.56} & \textbf{0.81} & \textbf{0.76} & 0.71\\
\hline
\end{tabular}
\end{center}
\end{table}
\setlength{\tabcolsep}{1.4pt}

\setlength{\tabcolsep}{4pt}
\begin{table}[!b]
\begin{center}
\caption{
RMS errors of recovered disparity for all pixels.
}
\label{tab:rmse_cmp}
\begin{tabular}{cccccccc}
\hline\noalign{\smallskip}
									& Buddha &Buddha2 &Horses &Medieval &Mona &Papillon &StillLife\\
\noalign{\smallskip}
\hline
\noalign{\smallskip}
GCDL\cite{wanner2012globally}     & 0.079 & 0.094 & 0.163 & 0.111 & 0.096 & 0.158 & 0.184\\
LAGC\cite{yu2013line}     			& 0.134 & 0.179 & 0.188 & 0.144 & 0.119 & 0.406 & 0.150\\
BCM\cite{chen2014light}				& \textbf{0.057} & 0.139 & 0.122 & 0.129 & 0.077 & \textbf{0.108} & 0.113\\
OADE\cite{wang2015occlusion}     	& 0.095 & 0.107 & 0.140 & 0.115 & 0.089 & 0.125 & 0.212\\
Our method 							& 0.069 & \textbf{0.051} & \textbf{0.074} & \textbf{0.101} & \textbf{0.071} & 0.148 & \textbf{0.110}\\
\hline
\end{tabular}
\end{center}
\end{table}
\setlength{\tabcolsep}{1.4pt}
\subsection{Depth maps}
\label{sec:ExpRe:dm}

The quantitative comparisons of the RMS errors of recovered disparity maps are listed in Tab. \ref{tab:rmse_cmp}. Note that all results are obtained in a same parameters setting. Our algorithm outperforms previous state-of-the-art algorithms in almost all datasets. 

\begin{figure}[!b]
\begin{center}
\centering
\subfigure[Buddha]{
	\label{fig:depthmap:buddha}
	\includegraphics[width=118mm]{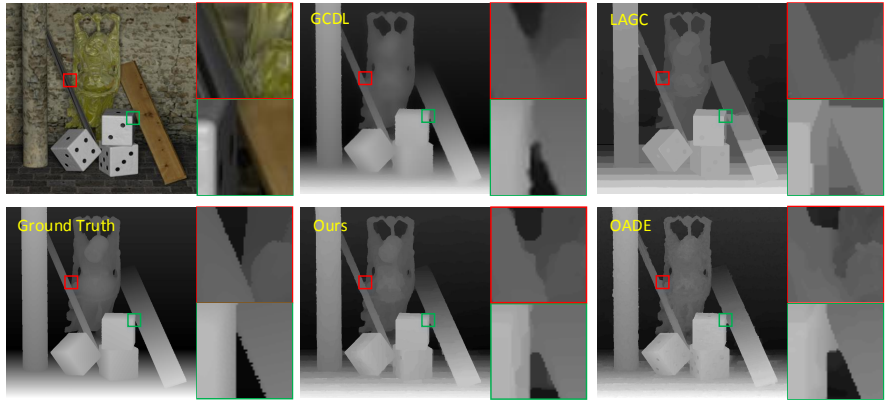} 
}
\subfigure[Buddha2]{
	\label{fig:depthmap:buddha2}
	\includegraphics[width=118mm]{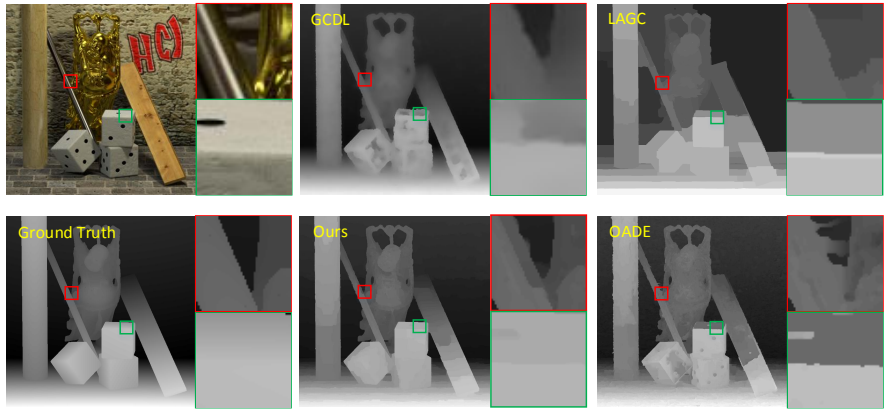} 
}
\end{center}
\caption{The disparity maps on the Buddha and Buddha2.}
\label{fig:depth_map1}
\end{figure}

\begin{figure}[!b]
\begin{center}
\centering
\subfigure[Mona]{
	\label{fig:depthmap:mona}
	\includegraphics[width=120mm]{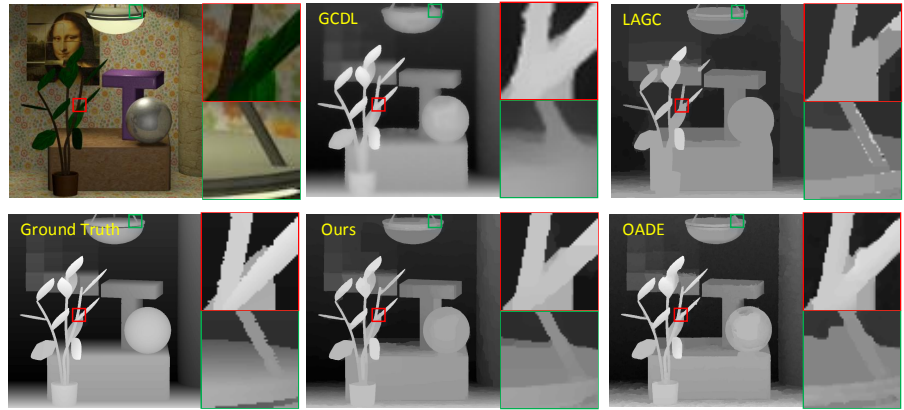} 
}
\subfigure[StillLife]{
	\label{fig:depthmap:stilllife}
	\includegraphics[width=120mm]{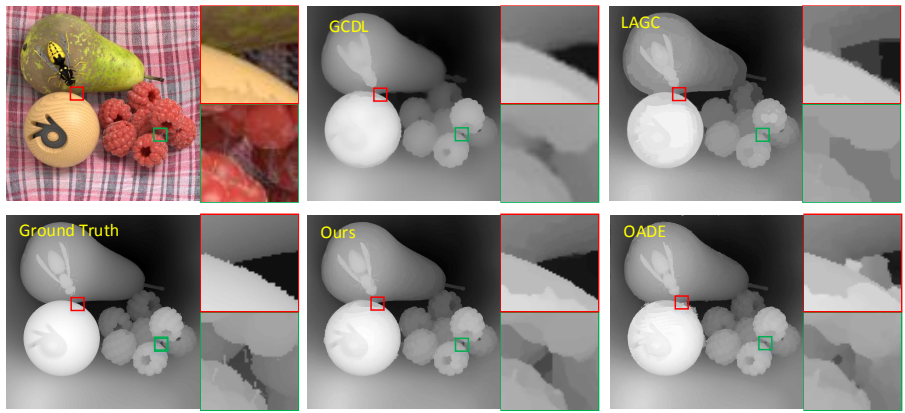} 
}
\subfigure[Horses]{
	\label{fig:depthmap:horses}
	\includegraphics[width=120mm]{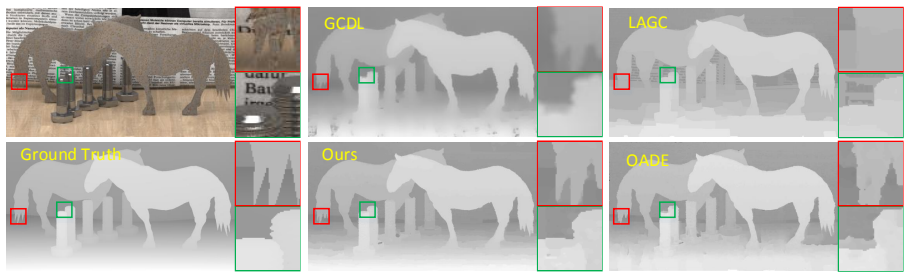} 
}
\end{center}
\caption{The disparity maps on the Mona, StillLife and Horses.}
\label{fig:depth_map2}
\end{figure}

The qualitative comparisons of the recovered disparity map are shown in Fig. \ref{fig:depth_map1}, \ref{fig:depth_map2}. It can be seen that, our algorithm yields sharper occlusion boundaries. As our selection method for un-occluded views can always find them accurately (Fig. \ref{fig:unocc_selection}, Tab. \ref{tab:oas_selection}), and do not select any other occluded views, our results are sharp in the multi-occluder areas. 

It is noticed the proposed algorithm performs not as good as OADE in the un-occluded views selection for Horses, however we get better results in depth map. This is because our energy function can preserve occlusion boundaries better. Moreover, we get the best results in the depth estimation for StillLife, however the F-measure of detected boundaries is not the best. The reason is that our algorithm performs much better than OADE in multi-occluder boundaries which have a larger difference of disparities.

Apart from the heavy occlusion, the proposed algorithm also performs well for shadings (Fig. \ref{fig:depthmap:buddha2}), although it is not taken into account in our model. Comparing with the Buddha (Fig. \ref{fig:depthmap:buddha}), there are more shadings in the buddha2 (Fig. \ref{fig:depthmap:buddha2}). Although all algorithms perform good in the Buddha, only our algorithm maintains the same level in the Buddha2. The reason for this phenomenon worth further study.

However, our algorithm can not handle the situation where the background has a similar texture or color to the occlusion. In the Fig. \ref{fig:depthmap:stilllife} (the green box), as the color of cloth in background is similar to the foreground, it is difficult to recover the true depth. Apart from this, our algorithm can not handle textureless areas just like all previous methods.

\section{Conclusion and Future works}
In this paper, we propose a new anti-occlusion depth estimation algorithm by modeling the formulation of the occlusion. The model reveals an important property, the occluders are consistent between the spatial and angular space. Utilizing this information, we improve the depth estimation in occlusion areas in two ways. Firstly, the un-occluded views are accurately selected by a clustering in spatial space, and the classical photo-consistency is enforced in these views. Secondly, the occlusion map is detected using the edges and the initial depth map, and then combined into the smooth term in the MRF function to keep the occlusion boundaries sharp. We have demonstrated the advantages of the proposed algorithm compared with other state-of-the-art algorithms in synthetic datasets.

Just as we mentioned in Sec. \ref{sec:ExpRe:dm}, our algorithm produces unexpected good results in shading situations although the shading is not considered in our model. It is worthy for us to investigate the phenomenon. Furthermore, as the light fields captured by real light field cameras have more noise compared with the synthetic datasets, it is essential to do more experiments on real data to better evaluate the performance of the proposed algorithm.

\bibliographystyle{plain}
\bibliography{egbib}

\end{document}